\documentclass[sigconf]{acmart}

\AtBeginDocument{%
  \providecommand\BibTeX{{%
    \normalfont B\kern-0.5em{\scshape i\kern-0.25em b}\kern-0.8em\TeX}}}

\usepackage[utf8x]{inputenc}
\usepackage{textcomp}

\setcopyright{none}
\renewcommand\footnotetextcopyrightpermission[1]{} 

\begin{document}

\title{Improvement in Machine Translation with Generative Adversarial Networks}

\author{Jay Ahn}
\authornote{All authors contributed equally to this research.}
\affiliation{%
  \institution{California Polytechnic State University}
  \city{San Luis Obispo}
  \state{California}
  \country{USA}
  \postcode{93407}
}
\email{jaahn@calpoly.edu}

\author{Hari Madhu}
\affiliation{%
  \institution{California Polytechnic State University}
  \city{San Luis Obispo}
  \state{California}
  \country{USA}}
\email{hmadhu@calpoly.edu}

\author{Viet Nguyen}
\affiliation{%
  \institution{California Polytechnic State University}
  \city{San Luis Obispo}
  \state{California}
  \country{USA}
}
\email{vnguy143@calpoly.edu}

\renewcommand{\shortauthors}{Ahn, Madhu, and Nguyen}

\begin{abstract}
In this paper, we explore machine translation improvement via Generative Adversarial Network (GAN) architecture. We take inspiration from RelGAN, a model for text generation, and NMT-GAN, an adversarial machine translation model, to implement a model that learns to transform awkward, non-fluent English sentences to fluent ones, while only being trained on monolingual corpora. We utilize a parameter $\lambda$ to control the amount of deviation from the input sentence, i.e. a trade-off between keeping the original tokens and modifying it to be more fluent. Our results improved upon phrase-based machine translation in some cases. Especially, GAN with a transformer generator shows some promising results. We suggests some directions for future works to build upon this proof-of-concept. 
\end{abstract}

\maketitle
\pagestyle{plain}

\section{Introduction}

Neural machine translation, the task of translating from one language (source) to another language (target) using a neural network, has gathered great interest in the deep learning community and models that have high accuracy have been developed over time, each outperforming the last\cite{bahdanau2016neural}\cite{vaswani2017attention} 

However, in practice, translated sentences can seem not fluent or not natural. While the choice of the translated words may be correct, it is rarely used by a native speaker in the given context. For example, the common greeting phrase "How are you doing?", when translated into Vietnamese by Google Translate, reads "Bạn đang làm gì?", which has the literal meaning of the original phrase rather than a question about someone's state-of-being. We do not know the architecture of Google's translation engine, but speculate that it is more simple and less accurate than the state-of-the-art systems in neural machine translation because of the number of users it serves.

Because a good deep learning model requires a lot of pairs of source and target languages, a more efficient solution could be training a model in the target language that can correct the mistakes the translator makes and produce a more fluent and natural utterance preserving the correct meaning of the input sentence. Having such a model would mean that we potentially only need a simple translator for each source language and use a single translation improvement model for the target language.

There are different ways to improve the fluency of a sentence: grammatical error correction, style transfer, or etc. We focused on improving the fluency of the sentence in a similar way of transferring style from text to text. In this paper, we introduce a generative adversarial net that improves the fluency of the translated sentences.

\section{Background}
\label{background}

Generative adversarial network is a novel deep learning architecture originally introduced in a field of computer vision. It corresponds to min-max two-player game between a generator and a discriminator where the discriminator tries to determine whether the sample is from the data distribution or model distribution, and the generator tries to fool the discriminator to determine its output to be valid. \cite{goodfellow2014generative} describes that the generative model can be thought of as analogous to a team of counterfeiters, trying to produce fake currency and use it without detection, while the discriminative model is analogous to the police, trying to detect the counterfeit currency.

However, applying the generative adversarial network to natural language text has the issue of the gradient update from the discriminator to the generator due to the non-differentiable argmax function from the output layer of the generator. Several approaches have been introduced to solve this issue such as REINFORCE algorithm and policy gradient\cite{yu2017seqgan} and Gumbel-softmax approximation \cite{kusner2016gans}.

In the field of natural language processing, the controlled generative adversarial network is widely used for style transfer, machine translation, or text summarization\cite{Betti2020ControlledTG, Ficler2017ControllingLS, hu2018controlled}. In the context of controlled/conditioned GAN \cite{mirza2014conditional}, original input in addition to the features as conditioning contexts gets fed to the generator and discriminator. For example, for style transfer, the characteristics of the style for target sentences is given to the generator and discriminator as input combined with the sentence or word embedding.

\section{Related Work}

\subsection{Monolingual Text to Text generation}
In the area of monolingual text to text generation, there have been approaches to transfer linguistic style, summarize the given text, or correct grammatical error using deep learning. Most of the recent work in style transfer or summarization have utilized a conditional language model, where the model gets trained with stylistic parameters or some knowledge of topic as conditioning contexts \cite{Singh2020IncorporatingSL, Ficler2017ControllingLS, Genest2011FrameworkFA}. In this paper, we don't improve the fluency of a text translated by a Google translator utilizing conditional language model, but we believe that the conditioned GAN is a interesting premise for future work.

\subsubsection{Grammar correction}
Grammatical error correction is one way to improve the awkwardness of translated language.
There have been very few relevant works for grammar correction using statistical modeling approaches. In \cite{ge2018reaching}, authors use a convolutional seq2seq model to learn sentence level fluency and how to infer/generate based on fluency learning. Sentence pairs are passed into the model, the fluency boost learning creates n-best inputs, which are scored and used as subsequent training instances. This way, the sentence is incrementally improved over a multi-round seq2seq inference and its fluency is improved. 

In \cite{omelianchuk2020gector}, a Grammatical Error Correction sequence tagger is proposed, with the intent of correcting and finding error tokens in a a given sentence. In this paper, the authors emphasize that sequence generation suffers from issues in inference speed, data availability, and interpretability so the task is simplified to sequence tagging. As such, authors propose multiple token-level transformations to a given sentence in order to continually correct it over multiple passes. In this system, as with the other grammar correction system, parallel-corpora of errorful and error-free data is used to train the model. 

However, in this paper, we assumed that the translated sentences are grammatically correct and focused on improving style of the sentence to look more natural.

\subsection{Applications of Generative Adversarial Networks}
Generative Adversarial Network has been widely applied to the area of natural language processing such as text generation, machine translation, and style transfer.

\subsubsection{Adversarial Text Generation}
As discussed in \nameref{background}, since the natural language is a discrete data, and the output comes from argmax of the probabilities from a softmax layer, text generation has the differentiability issue. Here, we discuss two major generative adversarial networks for text generation that achieve the state of the art performances: SeqGAN and RelGAN.

SeqGAN is a sequence generation generative adversarial net introduced by \cite{yu2017seqgan}. \cite{yu2017seqgan} fixed the generator differentiation problem by performing gradient policy update, modeling the data generator as a stochastic policy in reinforcement learning (RL). Like the standard generative adversarial network, its discriminator is trained with positive examples from data distribution and negative examples from the generator. However, unlike the standard one, the generator in SeqGAN gets updated by employing policy gradient and Monte Carlo search to consider the future outcome when generating the next token.

Another approach to overcoming the problem of sampling being a non-differentiable operation is to use the Gumbel-softmax activation \cite{jang2017categorical, maddison2017concrete}. This activation layer add noise from the Gumbel distributon, allowing for weighted sampling from the dictionary, and then use the softmax activation to pick the most likely token. This allows back-propagation of gradients from the discriminator to the generator, avoiding the use of a policy gradient algorithm.

According to \cite{nie2018relgan}, RelGAN performed better than other related frameworks like SeqGAN or LeakGAN\cite{guo2017long} developed on top of SeqGAN. Also, since it pretrains only the generator before adversarial training unlike others that pretrain both generator and discriminator, we adopted the RelGAN framework for our use case. In addition, we will use Gumbel-softmax to overcome the issue of non-differentiability in model sampling.

\subsubsection{Adversarial Neural Machine Translation}
The application of GAN to machine translation has been explored previously. In \cite{wu2018adversarial} and \cite{yang2018improving}, the authors use the conditional GAN framework, where the condition for both the generator and discriminator is the source language text. The generator learns to produce text in the target language that is indistinguishable by the discriminator from human-translated text. The discriminator learns to differentiate between machine-translated text and human-translated text given the source language text. In both of these papers, the authors followed SeqGAN's algorithm \cite{yu2017seqgan} to train the generator, which has an encoder-decoder architecture.

\subsubsection{Adversarial Style Transfer}
\cite{Lai2019MultipleTS} discusses a approach to generate sentences with a desired target style. Even though our work is based on RelGAN framework on the surface, we utilized the generator used in \cite{Lai2019MultipleTS} to get the sequence as an input.

\subsection{Machine Translation Detection}
\cite{Arase2013MachineTD} introduces a way to detect the machine translated sentences from monolingual web text. It utilizes some features extracted from human-generated sentences such as fluency features, grammatical features, gappy-phrase features, and so on. We believe that using these features as conditions in adversarial training would help the model to learn the features of natural sentences. But since this paper doesn't use conditioned GAN, it is included as a future work.

\section{System Architecture}
\begin{figure}
    \centering
    \includegraphics[width=\columnwidth]{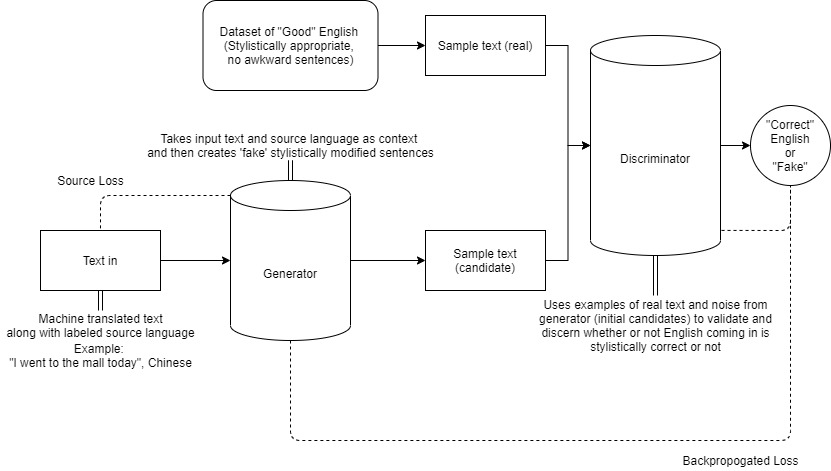}
    \caption{System architecture overview}
    \label{fig:system}
\end{figure}

An overview of our system architecture is shown in figure \ref{fig:system}. The model is split into two parts, a generator and a discriminator. There are two sources of data, one is the set of awkward English sentences that go into the generator, which is a sequence-to-sequence neural network that we hypothesize will transform them into fluent English sentences. The other source of data is fluent, stylistically proper English sentences. Together with the sentences generated by the generator, they are the input to the discriminator. The discriminator is a binary classification neural network that learns to classify between awkward and fluent English sentences. We label the fluent English sentences with the label 1, and the sentences from the generator with the label 0.

\subsection{Generator}
In this paper, we used two widely used sequence to sequence models for a generator: LSTM encoder-decoder and transformer in monolingual text to text generation \cite{Lai2019MultipleTS, Betti2020ControlledTG, Ficler2017ControllingLS}. 

\subsubsection{LSTM encoder-decoder}
\begin{figure}
    \centering
    \includegraphics[width=\columnwidth]{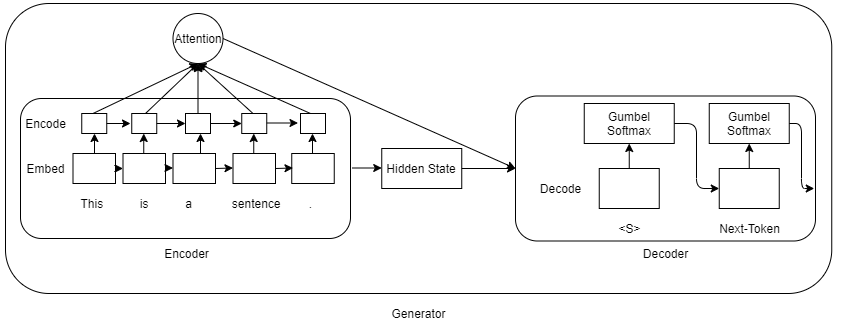}
    \caption{LSTM-based generator}
    \label{fig:lstm_seq2seq}
\end{figure}

We use a similar neural network architecture to Bahdanau et al.\cite{bahdanau2016neural}, shown in figure \ref{fig:lstm_seq2seq}
In each time step, the LSTM encoder receives a token of the input sequence and outputs a representation of that token together with past context. The LSTM layer also updates its context for the next time step. At the final time step, the context of the encoder, which represents the semantics of the input sentence, is passed to be the initial context of the decoder.

In each time step, the decoder receives a token that it has previously generated (initially a start token) and outputs a representation of the next token. Additionally, it also receives an attention context, a weighted sum of the outputs of the encoder, which is produced by a learned dense layer. This theoretically tells the decoder which tokens of the input sentence to focus on when generating the next token.

We modified the architecture of Bahdanau et al.\cite{bahdanau2016neural}, replacing the argmax layer on the decoder output with a Gumbel-softmax layer, so during training the decoder can sample from the vocabulary in a differentiable manner. During inference, we use a greedy strategy to produce a sequence, choosing the most likely token at each time step.

\subsubsection{Transformer}
\begin{figure}
    \centering
    \includegraphics[width=\columnwidth]{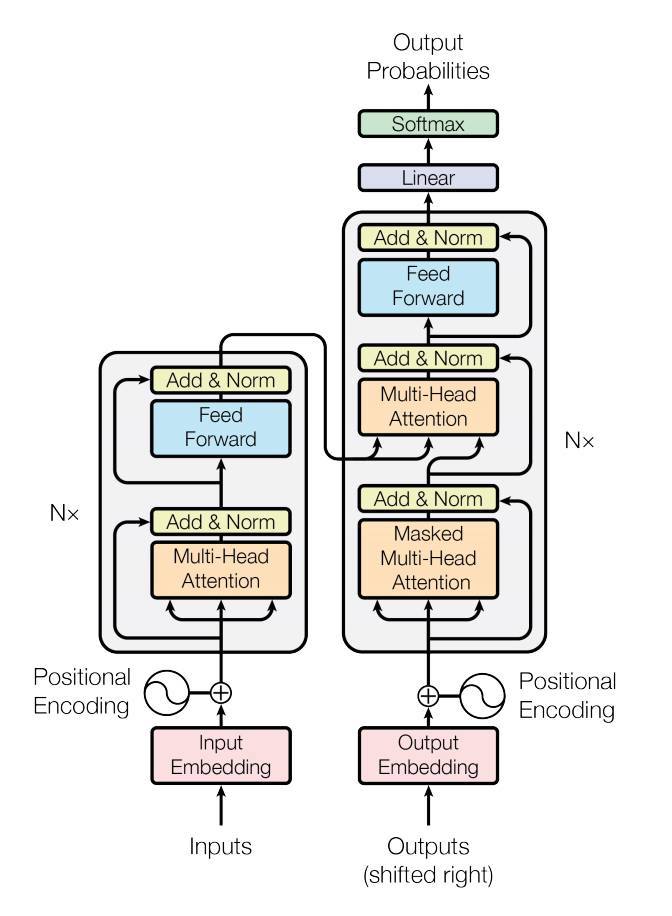}
    \caption{Transformer-based generator [18]}
    \label{fig:architure of transformer}
\end{figure}

The transformer architecture for a generator is same as the one first introduced in \cite{vaswani2017attention}. Figure \ref{fig:architure of transformer} describes the overall architecture of the transformer. Encoder consists of a block of multi-head attention with a normalization layer and a block of a point-wise feed forward network with a normalization layer. And, decoder contains two blocks of a multi-head attention with a normalization layer and a block of a point-wise feed forward network with a normalization layer.
While pretraining the generator, we use the regular transformer architecture described aforementioned with the sparse categorical cross entropy loss and Adam optimizer with the custom learning rate used in \cite{vaswani2017attention}. However, during adversarial training, we add a Gumbel softmax approximation layer\cite{nie2018relgan} on top of the decoder so that the generator loss calculated during the training can be back-propagated to the generator.
Refer to \cite{vaswani2017attention} for more details about how data flow within the architecture.

\subsection{Discriminator}
\begin{figure}
    \centering
    \includegraphics[width=\columnwidth]{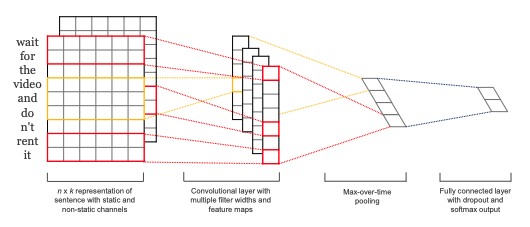}
    \caption{CNN Discriminator}
    \label{fig:CNN Discriminator}
\end{figure}

The discriminator is similar to the ones proposed by Yu et al.\cite{yu2017seqgan} and Nie et al.\cite{nie2018relgan}, as shown in figure \ref{fig:CNN Discriminator}. It includes multiple 1D convolutional layers, a flatten layer, and dense some more dense layers, and outputs a probability of the input sentence being a fluent English sentence.

\section{Training Setting}
\subsection{Data}

Our "good" styled sentences come from a translation of sentences in pre-cleaned Vietnamese Wikipedia pages that were open sourced on Tensor-Flow. The English translation of these sentences came from using Google's phase-based translation model on the aforementioned dataset of 80,000 Vietnamese sentences. This English translation was the input to our model, and what we then attempted to restyle with good structure and less awkwardness.

The translation was done using the Google Cloud API, and using an API key, where we were limited to generated translations for a maximum of 80,000 translation, which is why we had that limit for the number of sentences we could use. Getting our keys set up and able to generate was also another challenge we had to overcome, aside from the hard-limit that the API imposes on us.

\subsection{Training techniques}
Our goal is to train the model to a point where an equilibrium is reached: the discriminator's accuracy on inputs from the generator and fluent English inputs is 50\%, where it essentially guesses the outcome.

\subsection{Losses}
There are three main losses that we try to balance: the loss of the generator as an autoencoder ($L_AE$), the loss of the discriminator on inputs from the generator ($L_DG$), and the loss of the discriminator from fluent English sentence inputs ($L_DF$).

The loss of the generator as an autoencoder $L_AE$ is computed  as the categorical cross-entropy loss over each token of the input sequence against the corresponding token in the output sequence of the generator:

$L_AE = \sum_i^l{s_i * log(t_i)}$

where $s_i$ is the $i^{th}$ token of the input sentence, $t_i$ is the $i^{th}$ output token of the decoder of the generator, and $l$ is the number of tokens in the input sequence. The generator is trained to minimize this loss so that it learns a good language model.

As suggested by Goodfellow et al.\cite{goodfellow2014generative}, the loss of the discriminator on inputs from the generator $L_DG$ is computed as the log probability of the generator's output being a real English sentence:

$L_DG = log(D(G(s)))$

This is instead of the log probability of the generator's output not being a real English sentence $log(1 - D(G(s)))$ to make training the generator easier in the early stages \cite{goodfellow2014generative}. The generator is trained to maximize this loss, while the discriminator is trained to minimize it.

And similarly, the loss of the discriminator on inputs from the discriminator $L_DF$ is computed as the log probability of the real English sentence being a real English sentence:

$L_DF = log(D(x))$

The discriminator is trained to maximize this objective.

The general loss of the generator is then:

$L_G = L_AE - \lambda * L_DG$

were $\lambda$ is a hyperparameter that controls the relative importance of the two losses. Intuitively, we found that setting $\lambda$ very low encourages the generator to output the input sentence. However, while increasing it should theoretically encourage it to output a sentence similar to the real English sentences regardless of what the input sentence is, we observed that the generator generates seemingly random tokens.

\subsection{Training process}
As suggested by Yu et al. \cite{yu2017seqgan} and Nie et al. \cite{nie2018relgan}, we pretrain the generator as an autoencoder, i.e. only on minimize $L_AE$, for 20 epochs, before optimizing the other two losses. After this, we begin jointly training the generator and discriminator. We experimented with training them the same number of epochs, but found that the discriminator minimizes its loss very quickly at the beginning and the generator cannot improve its losses. Therefore, for every epoch the discriminator is trained, there are two epochs the generator is trained.

\section{Experiments}
For our evaluation data, we used two parallel corpora (Hindi-English and Vietnamese-English corpora) to have some measure of the ground-truth when finding a proper sentence to measure against. Both corpora feature hand-translated, stylistically-correct English that we can use for this purpose. The Vietnamese-English parallel sentences came from a corpus called EVBCorpus, which includes parallel sentence pairs mainly from news articles and film subtitles.The Hindi-English Corpus was taken from a data resource called CLARIN, and contains sentences from news articles, TED talks, and Wikipedia articles.

To test our model, we use two metrics: BLEU (Bilingual Evaluation Understudy Score) and sentence semantic similarity. The BLEU score is essentially the precision of the bag of words of a predicted sequence (the output of our model) against a ground truth translation:

$BLEU = \frac{|BOW(t) \cap BOW(r)|}{|BOW(t)|}$

where $t$ is the predicted sequence, $r$ is the ground truth sequence, and $BOW$ is a function that returns the set of unique tokens in a sequence. To calculate this, we use the built-in function $sentence_bleu()$ in the Python library NLTK.

While BLEU is widely used in machine translation, it has many shortcomings, such as not capturing the semantic similarity of words. For example, the words "car" and "automobile" can be interchangeable in many contexts, but to BLEU they are not. Thus we also utilize the cosine distance between the vector representations of the predicted sequence and the ground truth sequence. We use a pretrained BERT model to obtain these embeddings with the Python sent2vec library.

\section{Results}

\begin{table*}[]
    \centering
    \begin{tabular}{c}
\textbf{Given sentence}: The officials of the US. Virgin Islands and Puerto Rico said they are trying to restore the grid for more than 200,000 people.\\
\textbf{Model output}: the officials of the us . virgin islands and puerto rico said they are trying to restore the options for more than 200 , 000 people.\\
\hline
\textbf{Given sentence}: Families and communities facing the loss emotionally, physically and financially.\\
\textbf{Model output}: families and communities facing the loss to , physically and financially .\\
\hline
\textbf{Given sentence}: Many assert that their neighbors had seen Robert move back and forth from window to window, when the family is away.\\
\textbf{Model output}: many assert that their neighbors had seen robert move back and forth from navigate to carefully , when the family is away .\\
\hline
\textbf{Given sentence}:It is reported the development of VOA Special English program.\\
\textbf{Model output}: it is reported the development of voa special english program .\\
\hline
\textbf{Given sentence}: One is the increase in the number of women aged over 35.\\
\textbf{Model output}: one is the increase in the number of women aged over 35 .\\
\hline
\textbf{Given sentence}: Is going forward and how manipulation and how much cheating?\\
\textbf{Model output}: is going forward forward how manipulation and how much cheating ?\\
\hline
\textbf{Given sentence}: Is an incredible monument\\
\textbf{Model output}: is an month monument\\
\hline
\textbf{Given sentence}: I could not believe that I have your loyalty\\
\textbf{Model output}: i could not believe that i have your proposed\\
\hline
\textbf{Given sentence}: In keeping deep problems with the design students around the world,\\
\textbf{Model output}: in keeping problems with the design students students around the world ,
    \end{tabular}
    \caption{Output examples of the model}
    \label{tab:example_sentences}
\end{table*}

A few example outputs of the model are in table \ref{tab:example_sentences}. However, the most of the output from the adversarily trained generator is same as the input sentence as you can see in table \ref{tab:example_sentences}. We believe that this is the issue of a mode collapse (lack of diversity) or unstable generative adversarial training, and further research is required to improve the performance.

\begin{table}
    \centering
     \begin{tabular}{||c|c|c||} 
     \hline
     BLEU & Vietnamese-English & Hindi-English \\ [0.5ex] 
     \hline\hline
     Google Translate Baseline & 0.3461 & 0.3894 \\ 
     \hline
     GAN with LSTM Seq2Seq & 0.093 & 0.46 \\
     \hline
     GAN with Transformer  & 0.3416 & 0.409 \\
     \hline
    \end{tabular}
    \caption{The BLEU score on Vietnamese English parallel corpus and Hindi English parallel corpus}
    \label{tab:my_label}
\end{table}

As we can see with the above table, it would appear that the baseline already has some decent similarity to the ground-truth, although not perfect. With a BLEU score of around 0.3-0.4, there is some reasonable amount of similarity to a reference text. Interestingly enough, at least using the transformer, the GAN does a similar if not better job at attaining a score that is closer to reference. 

Notice that the GAN setup with the LSTM Seq2Seq architecture performs fairly poorly across all evaluations, but with the BLEU score of Hindi-English being almost artificially high compared to the rest of its performance. This is because there was a match between many unknown tokens between sentence embeddings that did, in fact, artificially raise the BLEU score only within the Hindi-English corpus to a reasonably high margin even though it is not truly close in meaning.

In contrast, the GAN using the Transformer does well largely due to the fact that it copies many if not all of the tokens that appear in the original baseline translation. This is why the scores as just as good if not better than the baseline translation here. The sentences are the same or just barely different to constitute a higher score, at least for the Hindi-English parallel corpus. These scores show that there is at least some real potential for the Transformer architecture, when trained with the right setup and hyper-parameter configuration, to help significantly in improving the quality of a given sentence.

\begin{table}
    \centering
    \begin{tabular}{||c|c|c||} 
     \hline
     Sent2Vec Similarity & Vietnamese-English & Hindi-English \\ [0.5ex] 
     \hline\hline
     Google Translate Baseline & 0.959 & 0.942 \\ 
     \hline
     GAN with LSTM Seq2Seq & 0.36 & 0.1 \\
     \hline
     GAN with Transformer  & 0.936 & 0.943 \\
     \hline
    \end{tabular}
    \caption{The Sent2Vec Similarity score on Vietnamese English parallel corpus and Hindi English parallel corpus}
    \label{tab:my_label}
\end{table}

With the Sent2Vec cosine similarity score, it is largely the same result as above. That is, the GAN with the LSTM architecture does quite poorly compared to baseline across the board due to inconsistencies and loss convergence issues in training time. At the same time, we see the GAN with the transformer architecture keep up at least reasonably well with the baseline translation, even with flaws in the training process. Similar to above, we believe that this shows the potential of the Transformer architecture GAN to perform better than the translation baseline with the right training process.

\section{Conclusion}
In this paper, we propose a GAN architecture off of RelGAN, the current state of the art text generation model with GAN. We tested two different GANs: one with an LSTM encoder-decoder generator and the other with a transformer generator. The GAN with the Transformer seems to work better than the one with LSTM sequence-to-sequence perhaps due to the fact that it is able to keep more context with multi-headed attention. 

In terms of BLEU, compared to the base line Google-translated sentences, GAN with transformer did not perform significantly poorly with Vietnamese to English parallel corpus, and it actually performed better with Hindi to English parallel corpus. Furthermore, in terms of sent2vec similarity to check semantic similarity, GAN with the transformer performed almost equally well as the baseline. Based on the results, a generative adversarial network to improve the awkwardness of the sentence is a promising direction to move forward.

\section{Future Work}

As we were quite new to adjusting to the training process of a GAN, we acknowledge that there are we are able to make in training and adjusting hyperparameters. For our model(s), we had somewhat naively adjusted our parameters, but did not achieve a way to automatically and quickly try and retry many parameter configurations for success criteria with regards to consistently dropping loss. 

Our model, as it stands currently, will work only with the one source language it is trained on, and will not be able to effectively handle multiple source languages. We would be able to improve this and allow the model to train on many different source languages by simply adding in a flag that denotes the source language to our input sequence.

Another flag that we would have been able to experiment with more was adding a tag that specifies the general grammar structure of the input sentence to give the model some ideas of common structures issues with certain sentences, and have a way to handle them separately.

Another aspect that we hope to improve in the future is quality and quantity of our input data. With access to maybe more API-keys per account, we would have to have access to more translated sentences, which is highly desired for giving our model more sentences to work with. In addition, we would also want some way to narrow down the domain of the sentences we use so as to use a more limited vocabulary in our model (the bigger the vocabulary, the larger the embedding layer and the more memory required to train).

Lastly, we attempted to look into changing the structure of the model to edit sequences at a world level instead of generating a completely new sequence with an encoder-decoder based model. In this setting, the input sentence would have been mostly preserved but some tokens could have been edited, deleted, or added to fix common stylistic issues. We attempted to use a base model similar to \cite{omelianchuk2020gector} for this task, but time did not allow for further exploration of this kind of model.

\bibliographystyle{ACM-Reference-Format}
\bibliography{sample-base}

\appendix

\end{document}